# A Dual Neighborhood Hypergraph Neural Network for Change Detection in VHR Remote Sensing Images

Junzheng Wu, Ruigang Fu, Qiang Liu, Weiping Ni, Kenan Cheng, Biao Li, Yuli Sun

**Abstract**

The very high spatial resolution (VHR) remote sensing images have been an extremely valuable source for monitoring changes occurred on the earth surface. However, precisely detecting relevant changes in VHR images still remains a challenge, due to the complexity of the relationships among ground objects. To address this limitation, a dual neighborhood hypergraph neural network is proposed in this article, which combines the multiscale superpixel segmentation and hypergraph convolution to model and exploit the complex relationships. First, the bi-temporal image pairs are segmented under two scales and fed to a pre-trained U-net to obtain node features by treating each object under the fine scale as a node. The dual neighborhood is then defined using the father-child and adjacent relationships of the segmented objects to construct the hypergraph, which permits models to represent the higher-order structured information far more complex than just pairwise relationships. The hypergraph convolutions are conducted on the constructed hypergraph to propagate the label information from a small amount of labeled nodes to the other unlabeled ones by the node-edge-node transform. Moreover, to alleviate the problem of imbalanced sample, the focal loss function is adopted to train the hypergraph neural network. The experimental results on optical, SAR and heterogeneous optical/SAR data sets demonstrate that the proposed method comprises better effectiveness and robustness compared to many state-of-the-art methods.

## 1. Introduction

As a vital branch of remote sensing interpretation, change detection (CD) that aims at identifying the changes in images collected in the same area but at different times (Singh, 2010) has been widely applied in various fields, including urban planning, damage assessments, environmental monitoring (Canuti et al., 2004). With the rapid developments of Earth observation technology, very high spatial resolution (VHR) remote sensing images are now available, and they can provide abundant surface details and spatial distribution information. Consequently, CD in VHR images has drawn more attention and meanwhile, is a tough and challenging task (Huang et al., 2020).

In the last few decades, numerous CD methods have been developed for diverse circumstances, and they can be divided into pixel-based and object-based methods roughly according to the basic unit of processing (Hussain et al., 2013). The pixel-based CD (PBCD) methods employ individual pixel as the basic analysis unit, and many widely used methods belong to this category. For optical images CD tasks, change vector analysis (CVA) (Bruzzone and Prieto, 2000) and its extensions, such as robust CVA (RCVA) (Thonfeld et al., 2016) are the most commonly used ones, which can provide change intensity and change direction. Some transform-based method, such as principal component analysis (PCA) (Celik, 2009) and slow feature analysis (SFA) (Wu et al., 2014) are also popular. To suppress the interference of speckle noise, ratio-based approaches (Bovolo and Bruzzone, 2005; Inglada and Mercier, 2005) were devised for SAR CD. Some statistics-based methods (Yang et al., 2019) used specific distributions to model SAR images and then obtained change maps through estimating the posteriori probabilities. Despite fewer comparing with those for homogeneous CD, quite a few methods have emerged in heterogeneous optical/SAR CD tasks. For instance, Mercier et al (2008) transformed one of the two images to the other one using the copula theory, then, the Kullkack–Leibler (KL) distance was employed to calculate the change indices. Ferraris et al (2019) used the coupled dictionary learning framework to model the two heterogeneous images. To capture the information of the adjacent pixels, numerous methods used a sliding window to obtain patches as processing units (Chatelain et al., 2008). In addition, various advanced techniques have been introduced into PBCD, such as Markov random fields (Chen and Cao, 2013), extreme learning machine (ELM) (Chang et al., 2010) and dictionary learning (Gong et al., 2016). However, These PBCD methods are limited on VHR images, because the assumption of pixel independence is unreliable. When using a sliding window, it is challenging to obtain the most appropriate size. Besides, a fixed window can hardly represent a meaningful ground object. Furthermore, registration error and radiometric correction have a significant influence on the results (Zhang et al., 2021).

With the spatial resolution increasing, the highly spectral variability and the difficulty of modeling the contextual information may further weaken the performances of PBCD methods. Under these circumstances, object-based CD (OBCD) techniques provide a unique way to overcome these limitations by using an object as a basic processing unit. An object (so-called parcel or superpixel in some literature) is a group of local pixel clusters that are obtained through segmentation using spectral, texture, geometric (such as shape) feature, and other information. Large numbers of OBCD methods have been proposed in the last decade. For instance, Yousif and Ban (2017) developed a novel OBCD method for VHR SAR images, which can preserve meaningful detailed change information and mitigate the influence of noise. Multiple feature fusion strategy was designed to improve the performance of OBCD in (Wang et al., 2018). Lv et al (2020) proposed an object-oriented key point vector distance to measure the changed degree for VHR images, which can reduce the number of pseudo changed points. These methods have shown extraordinary potential to weaken the effects of spectral variability, spatial georeferencing, and acquisition characteristics (Zhan et al., 2020). Nevertheless, most OBCD methods only use some hand-crafted features, which require much domain-specific knowledge and may be affected by noise and atmospheric conditions. Furthermore, to ensure the homogeneity within each object, most OBCD methods make the objects prone to over-segmentation, and the boundary fragmentation may lead to inadequately semantic integrity. It implies that the complex interactions among objects need to be further exploit and this is a vital motivation of our method.

Recently, deep learning techniques have demonstrated remarkable performance in image processing field due to their capabilities of automatically obtaining abstract high level representations by gradually aggregating the low-level feature, by which the complicated feature engineering can be avoided (Ball et al., 2017), and without doubt, various deep learning methods have been employed in CD tasks, including both supervised/ semi-supervised and unsupervised manners. Among these deep learning methods, convolutional neural networks (CNN) has drawn intensive attention and has been the most popularly used as the backbone or feature extractors in CD tasks. For instance, Lim et al (2018) designed three encoder-decoder structured CNNs to yield change maps using Google Earth images. Chen and Shi (2020) presented a novel spatial-temporal attention neural network based on Siamese and designed a CD self-attention mechanism to calculate attention weights between any two pixels at various times and positions. Wu et al (2021) proposed a deep kernel PCA convolution to extract representative features for multitemporal VHR images, then, built an unsupervised Siamese mapping network for binary and multiclass CD. An end-to-end CD method that combines a pixel-based convolutional network and superpixel-based operations was proposed in (Zhang et al., 2021), which promoted the collaborative integration of OBCD and deep learning. Wang et al (2021) designed a fully convolutional Siamese network and improved the focal contrastive loss, which can reduce intra-class variance and enlarge inter-class difference. Several generative adversarial network (GAN) architectures based on CNN units have been also exploited for CD (Gong et al., 2017; Liu et al., 2020). Although the exiting deep learning based methods have achieved promising performances in some cases, some limitations still exist. To be specific, ground objects in RS images commonly appear with various shapes, whereas the CNN models only conduct the convolution on the regular rectangular regions. In other works, the geometric information of objects cannot be completely captured by CNN. Besides, the weights of each convolution kernel are identical when convolving all patches. As a result, the boundaries between changed and unchanged classes may be lost. In addition, CNNs capture abstract information through sequential down-sampling operation, which may lead to inadequately exploiting the interactive relationships among objects.

In view of the aforementioned issues, we propose an object-level semi-supervised hypergraph neural network (HGNN) framework for VHR image CD tasks in this article, acting upon both homogeneous and heterogeneous remote sensing images. The input images are firstly segmented under a fine scale parameter and a coarse one, respectively. Treating each object under the fine scale as one node, the high-level features of node are obtained by combing segmentation under the fine scale and the outputs of the feature extractor, which is a pre-trained U-net. The hypergraph is then constructed by defining the dual neighborhood to obtain reliable hyperedges, with the goal of capturing more comprehensively structural information among objects. Several hypergraph convolution layers are then sequentially conducted on the hypergraph to propagate information form a small amount of labeled nodes to the unlabeled ones through the hypergraph structure. As changed and unchanged samples as usually imbalanced in CD, the focal loss function is adopted when training the network.

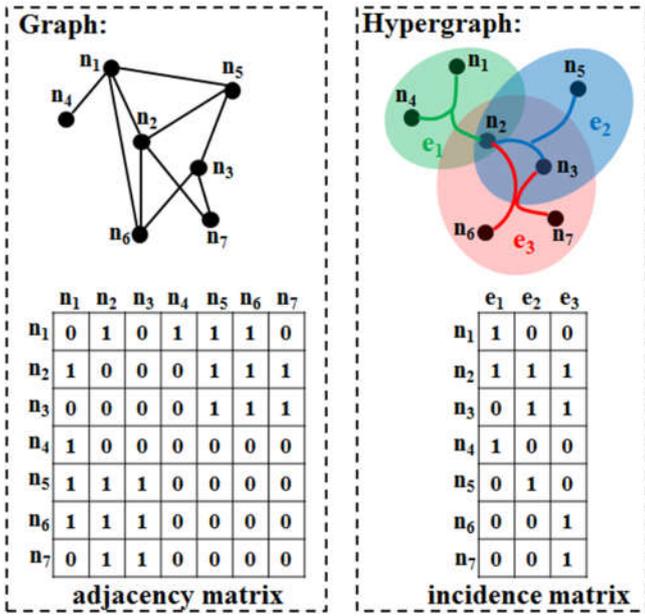

Fig.1. An example of ordinary graph and hypergraph

The main contributions of this article are as follows.

1) We propose a novel dual neighborhood hypergraph neural network (DNHGNN) framework for CD, which can adequately exploit the complex relationships and interacting information of ground objects that commonly exist in the VHR remote sensing images. To the best of our knowledge, it's the first HGNN based method in the field of remote sensing CD.

2) A dual neighborhood is defined, which contains a spatial neighborhood according to the adjacent relationships of object under the fine scale and a structural neighborhood according to the father-child relationships between scales. Based on the dual neighborhood, reliable hyperedges can be obtained, which better represent the complicated structures of images.

3) The multiscale object-based technique is integrated into hypergraph construction, which not only yields high-level node features as inputs of HGNN, but also substantially reduces the number of nodes, making the hypergraph convolution on the image scene feasible and efficient.

## 2. Related Works

In this section, we provide a brief review of graph/ hypergraph based CD methods and graph neural networks, as they are related to this article.

### 2.1 Graph/Hypergraph Based Change Detection

Fig. 1 shows an example of ordinary graph and hypergraph. The main difference is that the graph is represented using the adjacency matrix, in which each edge connects just two vertices, while the hypergraph can be represented by the incidence matrix, in which the hyperedges can encoder high order correlations by connecting more than two vertices.

The VHR images are usually full of structural feature, and the graph/hypergraph models are feasible in representing these features. Thus, they have been utilized in CD. Pham et al (2016) proposed a weighted graph to capture contextual information of a set of characteristic points and measured the changed level by considering the coherence of the information carried by the two images. A hierarchical spatial-temporal graph kernel was designed to utilize the local and global structures for SAR images CD (Jia et al., 2020). An object-based graph model is proposed for both optical and SAR images CD in (Wu et al., 2021). Sun et al (2021) constructed a nonlocal patch similarity graph to establish a connection between heterogeneous images, and then determined the change level by measuring how much the graph structure of on image still conforms to that of the other image. Despite much fewer comparing with graph based methods, several works have attempted to introduce hypergraph models into CD tasks. For instance, Wang et al (2020) formulated the CD as the problem of hypergraph matching and hypergraph partition, and constructed the hyperedges on the pixels and their coupling neighbors using the $K$ nearest neighbors rule.

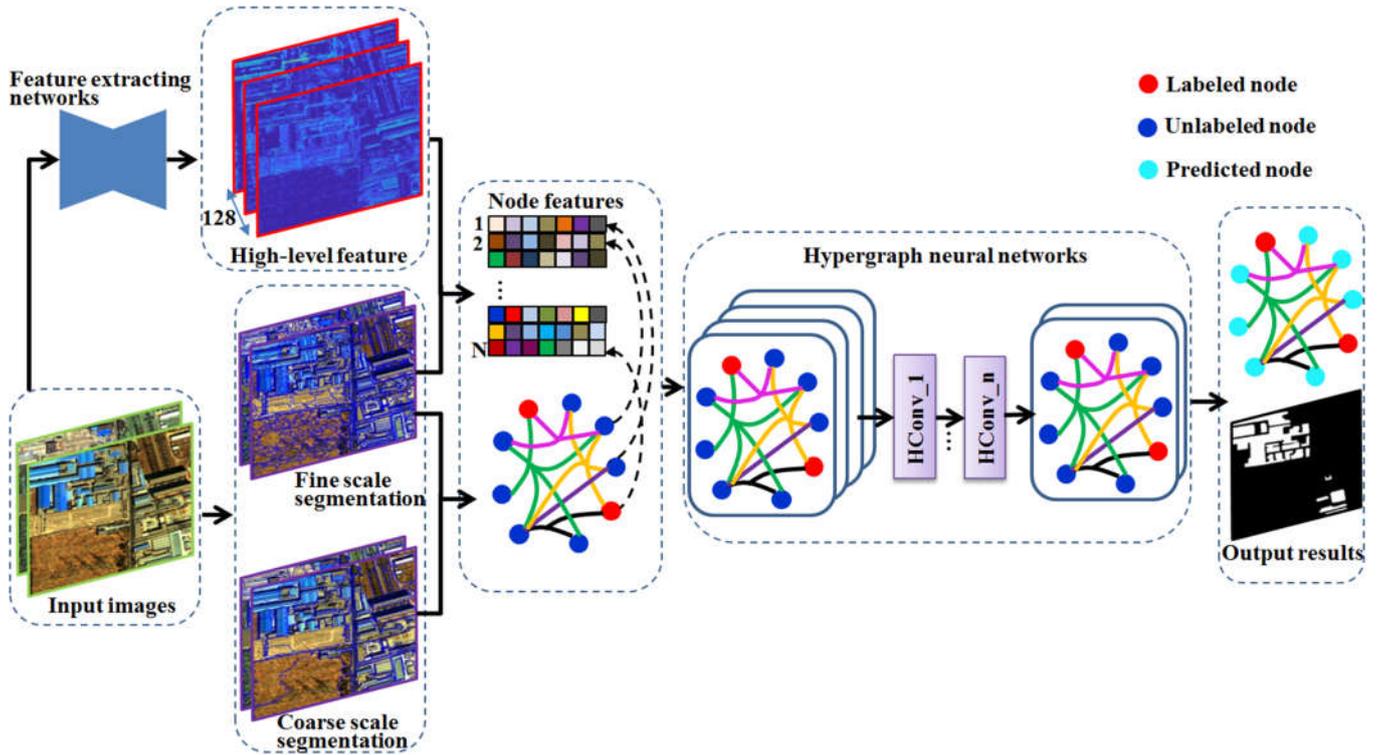

Fig.2. Flowchart of the proposed method.

Although the aforementioned graph/hypergraph based methods have validated the availability in many cases, some limitations still exist. To be specific, they only utilize some low features (e.g. spectrum, intensity, texture), which may not only be affected by noise and atmospheric conditions, but also contain insufficiently semantic information. In addition, the complexly structural relationships in high spatial resolution images have not been fully exploited when constructing the graph/hypergraph models.

**2.2 Graph Convolutional Network**

The notion of graph neural network (GNN) was initially outlined in 2005 (Gori et al., 2005). Wu et al (2021) have made a comprehensive survey on GNN. Motivated by CNNs, RNNs, and autoencoders form deep learning, new generalizations and definitions of important operations have been rapidly developed over the past few years. The most representative one is graph convolutional networks (GCN), which can be divided into two streams: the spectral-based approaches and the spatial-based approaches. In the spectral-based stream, Bruna et al (2014) used a spectral convolution to define a multilayer neural network model, which is similar to the classical CNN. Up to date, there have been numerous extensions on spectral-based approaches (Defferrard et al., 2016; Levie et al., 2019). The research on spatial-based stream started much earlier. Micheli (2009) first addressed mutual dependence using a spatial-based approach. Kipf and Welling (2017) proposed a fast approximation localized convolution and designed a simple layer-wise propagation rule for semi-supervised classification, which makes the GCN model able to encode both graph structure and node features. Since then, increasing extensions and improvements have emerged, such as FsatGCN (Chen et al., 2018) and the autoencoder-constrained graph convolutional network (AEGCN) (Ma et al., 2021). With the capability of modeling the irregular data structures, GCN has been widely applied to various vision tasks, such as specific object detection (Yan et al., 2019) and hyperspectral image classification (Wan et al., 2020). In the recent two years, several CD methods based on GCN have emerged. For instance, Saha et al (2020) first utilized GCN for semi-supervised CD tasks. Tang et al (2021) combined GCN and the metric learning to develop an unsupervised method for CD. In our previous works (Wu et al., 2021), a multiscale GCN for CD was proposed, which can comprehensively incorporate the information from several distinct scales.

In traditional GNN, the pairwise relationships are employed, as each edge connects only two vertices. However, in many real applications, the object relationships are much more complex than pairwise, such as social connections and recommendation systems. Under such circumstances, traditional GNN has intrinsic limitations in modeling these relationships. To tackle this challenging issue, Feng et al (2019) recently proposed the hypergraph neural networks (HGNN), which used the hypergraph structure for data model, then, a hypergraph convolution operation was designed to better exploit the high order data correlation for representation learning. An improvement named dynamic hypergraph neural networks (DHGNN) was proposed, which modified the hypergraph structure from adjusted feature embedding (Jiang et al., 2019).

Based on the observations that relationships of ground objects in VHR remote sensing images may be beyond pairwise connections and even far more complicated, in this article, we propose a HGNN framework for VHR image CD, which aims at improving the performances by better exploiting the complex relationships among objects.

### 3. Methodology

In this section, the proposed method is introduced in details. The flowchart can be seen in Fig.2. Firstly, the two input images are stacked into one image and then segmented under a fine scale parameter and a coarse one, respectively. Meanwhile, the input images are fed into the feature extractor which is a pre-trained U-net, to produce the high-level feature maps with the same lengths and heights as input. After that, the object-wise features of nodes are obtained by combining the segmentation under the fine scale and the high-level feature maps. While the hypergraph can be constructed using the adjacent relationships under the fine scale and the father-child relationships between the two scales. Then, sequential hypergraph convolutional layers are conducted on the hypergraph, which exploit the high order relationships carried by hypergraph structure to cluster the objects potentially belonging to the same class (changed/unchanged) together in the embedding space.

**3.1 Multiscale Segmentation**

In order to obtain homogeneous regions (namely objects) as the basic processing units which contain sufficient spatial information, object-based image analysis always starts with object segmentation, a process of partitioning an image into small separate regions (segments) according to certain criteria. We adopt fractal net evolution approach (FNEA) (Baatz and Schape, 2000; Yang et al., 2014) for the segmentation task, which is a bottom-up region merging technique with a fractal iterative heuristic optimization procedure. The objects are obtained by merging regions according to an optimization function, which requires the heterogeneity of the merged object in terms of spectral and shape properties to be lower than the user-defined threshold. The FNEA must be triggered by three parameters, namely, *scale*, *shape*, and *compactness* during the region merging. The *scale* parameter decides the maximum allowed heterogeneity for image objects, the *shape* parameter determines the aim of the entire homogeneity criterion, and *compactness* is used to optimize image objects. More details of FNEA can be tracked in (Baatz and Schape, 2000; Yang et al., 2014).

Compared with other segmentation approaches such as SLIC and mean-shift, FNEA is adopted in this article based on the following advantages. First, FNEA can better produce boundary-constrained multiscale segmentation [54]. Second, not only spectral properties but also geometric information are taken into account during segmentation, thus, objects with various shapes can be extracted with relatively high accuracy [55]. Third and foremost, when keeping the *shape* and *compactness* parameters invariant, father-child relationships exist between segmented scales. Specifically, each object under a coarse scale parameter is composed of several objects under a finer one, and these relationships actually contain abundantly structural information of ground objects that can be exploited to construct hypergraph.

**3.2 Feature Extracting Network**

In many vision tasks, feature extraction and selection is a complex but rather pivotal step that requires professional knowledge and experience. Considering that the improved version of fully convolutional neural network (FCN) called U-net has been effectively exploited to some CD works (Daudt et al., 2018; Liu et al., 2020), we adopt and modify the original U-net structure to extract pixel-wise high level features, which could avoid the hand-crafted designs for extraction and deficient modeling of the features.

As shown in Fig.3, the network consists three parts that involve a contracting (or encoding) path, bottleneck, and expanding (or decoding) paths. The details of the network can be tracked in our previous work (Wu et al., 2021). The cross entropy function is adopted as the loss function with the following formula:

$$loss = -\sum_{n}\left[y_n \ln a_n + (1-y_n)\ln(1-a_n)\right] \quad (1)$$

where $y_n$ is the real value of the sample, and $a_n$ is the actual output result.

In this article, the feature extracting network is pre-trained using the openly free ONERA Satellite Change Detection OSCD

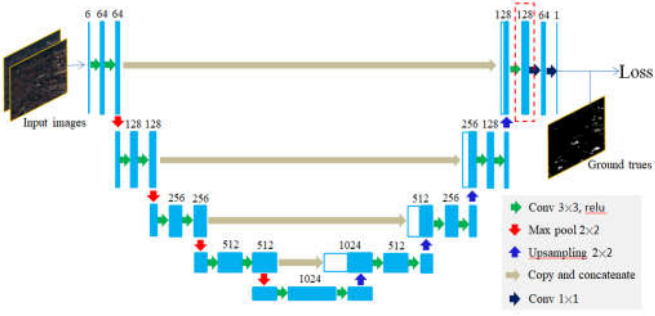

Fig.3. Illustration of feature extracting network

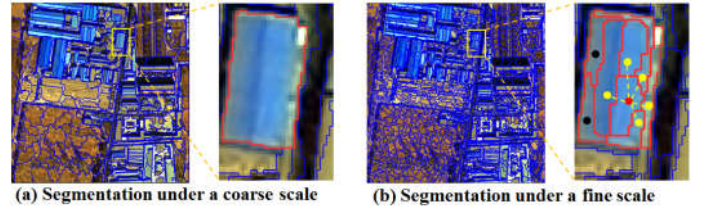

Fig.4. Illustration of dual neighborhood

dataset (Daudt et al., 2018). The input images are clipped into patches with the size of 112×112 in the pre-training process. When the network has been trained, it can be used as the feature extracting network as the module shown in Fig.2, and the input images can be free-size without any adjustment. It is worth noting that the 128 feature maps (marked by the red box in Fig.3) of the last convolution operations set are treated as the extracting results of the feature extracting network and fed to the following hypergraph construction, and we denote the feature maps as $\mathbf{M} \in \mathbf{R}^{H \times W \times 128}$, where $H$ and $W$ are the height and width of the input images, respectively.

### 3.3 Hypergraph Construction

Supposing that the input images $I_1, I_2$ are stacked into one image $I_S$, then, $I_S$ is segmented by the FNEA under a fine scale parameter $S_1$ and a coarse one $S_2$ with the same of *shape* and *compactness* parameters, respectively. After that, two object sets can be obtained: $\Omega_1 = \{F_1, F_2, \cdots, F_N\}$ and $\Omega_2 = \{C_1, C_2, \cdots, C_M\}$, $F_i$ and $C_j$ ($i=1,2,\cdots,N; j=1,2,\cdots M$) denote the objects under the fine and coarse parameters, respectively, and naturally $N > M$. In this article, we treat each object under the fine scale parameter $S_1$ as a vertex, thus, a hypergraph can be defined as $\mathbf{G} = (\mathbf{V}, \varepsilon, w)$, which includes a vertex (node) set $\mathbf{V} = [\mathbf{v}_1, \cdots, \mathbf{v}_N]^T \in \mathbf{R}^{N \times d}$, a hyperedge set $\varepsilon = [\mathbf{E}_1, \cdots, \mathbf{E}_L]$, and each hyperedge $\mathbf{E} \in \varepsilon$ is assigned a positive weight $w(\mathbf{E})$, where $d$ is the length of each feature vector, $d=128$.

As $\mathbf{M} \in \mathbf{R}^{H \times W \times 128}$, the feature maps of input images have been obtained through the pre-trained U-net, to build the object-wise feature $\mathbf{v}_i \in \mathbf{R}^{1 \times 128}$ representing the *i*th node, we combine the $\mathbf{M} \in \mathbf{R}^{H \times W \times 128}$ which can represent high level semantic information and the object set $\Omega_1 = \{F_1, F_2, \cdots, F_N\}$ that contain abundantly spatial information to characterize the node. $\mathbf{v}_i$ is formulated by:

$$\mathbf{v}_i = \sum_{\{(j,k) | I_S(j,k) \in F_i\}} \mathbf{M}(j,k,:) / \#(F_i) \tag{2}$$

where $\#(F_i)$ is the number of pixels in $F_i$.

Most existing approaches of constructing hyperedge treat a centroid vertex with its K-nearest neighbors (KNN) in the spatial domain, or the centroid vertex and all vertices adjacent to it as the vertices within a hyperedge. The former may suffer the limitation that a large K would cause remarkable redundancy, while a small K would acquire deficient local relationships. Besides, a fixed K can't be appropriate for all regions in VHR images due to the diversity of ground objects. The latter is based on the widely used observation that the adjacent vertices (a vertex can present a pixel, a patch, an object and so on) have a greater probability to be the same class. However, using only the adjacent vertices may be insufficient in some cases. Consequently, we define the dual neighborhood to construct informative hyperedges which can better exploit the relationships among objects.

As illuminated in Fig.4, a building is presented by only one object (region) under the coarse scale, while it is over-segmented into many objects under the fine scale (remarked by red boundaries). Supposing the red solid node in Fig.4 (b) is the centroid vertex, if the hyperedge is constructed only using itself and the adjacent vertices (yellow solid nodes), the structure of building may not be fully represented. It can be intuitively observed that the objects denoted by black solid nodes are not adjacent to the centroid one, they still prone to belong to the same ground object, due to they are all children of the same object under the coarse scale parameter. Thus, when constructing the hyperedge according to the red solid nodes, the structural information can be better exploited if taking both yellow and black solid nodes into account.

According to the analysis above, we construct a hyperedge for each $\mathbf{v}_i$ based on the spatial neighborhood and structural neighborhood.

The spatial neighborhood of $\mathbf{v}_i$ is defined as follow:

$$N_1(\mathbf{v}_i) = \{\mathbf{v}_j \mid F_i, F_j \in \Omega_1, F_j \text{ is adacent to } F_i\} \tag{3}$$

While the structural neighborhood of $\mathbf{v}_i$ is defined as follow:

$$N_2(\mathbf{v}_i) = \{\mathbf{v}_j \mid F_i \in C_k \ \& \ F_j \in C_k\} \tag{4}$$

Consequently, the hyperedge $\mathbf{E}_i$ centers on $\mathbf{v}_i$ is denoted

as:
$$\mathbf{E}_i = \mathbf{v}_i \cup N_1(\mathbf{v}_i) \cup N_2(\mathbf{v}_i) \quad (5)$$

The incidence matrix $\mathbf{H} \in \mathbf{R}^{N \times L}$ ($L=N$ in this article) of the hypergraph is represented as:

$$\mathbf{H}(i,j) = h(\mathbf{v}_i, \mathbf{E}_j) = \begin{cases} 1, \text{if } \mathbf{v}_i \in \mathbf{E}_j \\ 0, \text{otherwise} \end{cases} \quad (6)$$

The hyperedge weight is computed as:

$$w_i = w(\mathbf{E}_i) = \frac{2}{n_i(n_i-1)} \sum_{j,k|\mathbf{v}_j \in \mathbf{E}_i, \mathbf{v}_k \in \mathbf{E}_i} \exp(-\|F_j - F_k\|) \quad (7)$$

where $n_i$ is the number of vertices within $\mathbf{E}_i$. $w_i$ measures the similarity of all vertices within the hyperedge. A large $w_i$ means that the similar degree is high, thus the vertices within the hyperedge are prone to belong to the same class.

Based on $\mathbf{H}$ and $w$, the vertex degree of each $\mathbf{v}_i \in \mathbf{V}$ is:

$$d_i = d(\mathbf{v}_i) = \sum_{j=1}^{N} w_j \mathbf{H}(i,j) \quad (8)$$

and the edge degree of each $\mathbf{E}_i \in \varepsilon$ is:

$$\delta_i = \delta(\mathbf{E}_i) = \sum_{j=1}^{N} \mathbf{H}(j,i) \quad (9)$$

Further, $\mathbf{W} = diag(w_1, \cdots, w_N)$, $\mathbf{D}_\mathbf{v} = diag(d_1, \cdots d_N)$ and $\mathbf{D}_\mathbf{E} = diag(\delta_1, \cdots, \delta_N)$ denote the diagonal matrices of the hyperedge weights, the edge degrees and the vertex degrees, respectively.

Then, the normalized hypergraph Laplacian matrix can be computed as follow:

$$\Delta = \mathbf{I} - \mathbf{D}_\mathbf{v}^{-1/2} \mathbf{H} \mathbf{W} \mathbf{D}_\mathbf{E}^{-1} \mathbf{H}^T \mathbf{D}_\mathbf{v}^{-1/2} \quad (10)$$

where $\Delta$ is positive semi-definite.

### 3.4 Hypergraph Convolution

Given a signal $\mathbf{x} = (\mathbf{x}_1, \cdots, \mathbf{x}_n)$ with hypergraph structure, the approximation of spectral convolution of $\mathbf{x}$ and filter $\mathbf{g}$ can be denoted as (Defferrard et al., 2016):

$$\mathbf{g} * \mathbf{x} \approx \sum_{k=0}^{K} \theta_k T_k(\tilde{\Delta}) \mathbf{x} \quad (11)$$

where $T_k(\tilde{\Delta})$ is the Chebyshev polynomial of order $k$ with scaled Laplacian $\tilde{\Delta} = \frac{2}{\lambda_{max}} \Delta - \mathbf{I}$, with $\lambda_{max}$ being the largest eigenvalue of $\Delta$. Equation (11) can avoid the expansive computation of Laplacian eigenvectors by using only matrix powers, additions and multiplications. Feng et al (2019) further let $K=1$ to limit the order of convolution operation for the reason that the Laplacian in hypergraph can already well represent the high order relationships among nodes. Kipf and Welling (2017) pointed out that $\lambda_{max} \approx 2$ as the neural network parameters can adapt in scale during the training process. Based on the above assumptions, the convolution operation can be simplified to:

$$\mathbf{g} * \mathbf{x} \approx \theta_0 \mathbf{x} - \theta_1 \mathbf{D}_\mathbf{v}^{-1/2} \mathbf{H} \mathbf{W} \mathbf{D}_\mathbf{E}^{-1} \mathbf{H}^T \mathbf{D}_\mathbf{v}^{-1/2} \mathbf{x} \quad (12)$$

where $\theta_0$ and $\theta_1$ is parameters of filters over all nodes. Since it can be beneficial to constrain the number of parameters to address over fitting and to minimize the number of operations (such as the matrix multiplications), by setting $\theta_0$ and $\theta_1$ as:

$$\begin{cases} \theta_1 = -\frac{1}{2}\theta \\ \theta_0 = \frac{1}{2}\theta \mathbf{D}_\mathbf{v}^{-1/2} \mathbf{H} \mathbf{D}_\mathbf{E}^{-1} \mathbf{H}^T \mathbf{D}_\mathbf{v}^{-1/2} \end{cases} \quad (13)$$

Equation (12) can be simplified to the following expression:

$$\mathbf{g} * \mathbf{x} \approx \theta \mathbf{D}_\mathbf{v}^{-1/2} \mathbf{H} \mathbf{W} \mathbf{D}_\mathbf{E}^{-1} \mathbf{H}^T \mathbf{D}_\mathbf{v}^{-1/2} \mathbf{x} \quad (14)$$

Then, given a hypergraph signal $\mathbf{X} \in \mathbf{R}^{N \times I_1}$ with $N$ nodes and $I_1$ dimensional feature, the hypergraph convolution can be formulated by

$$\mathbf{Y} = \mathbf{D}_\mathbf{v}^{-1/2} \mathbf{H} \mathbf{W} \mathbf{D}_\mathbf{E}^{-1} \mathbf{H}^T \mathbf{D}_\mathbf{v}^{-1/2} \mathbf{X} \Theta \quad (15)$$

where $\Theta \in \mathbf{R}^{I_1 \times I_2}$ is the parameter to be learned during the training process. The output $\mathbf{Y} \in \mathbf{R}^{N \times I_2}$ can be used for clustering or classification.

Noting that $\mathbf{D}_\mathbf{v}$ and $\mathbf{D}_\mathbf{E}$ play a role of normalization in equation (15), the hypergraph convolution actually propagates information by a node-edge-node manner, which can better refine the features using the hypergraph structure. Equation (15) can be interpreted from right to left as follows: first, the initial node feature $\mathbf{X}$ is processed by the filter matrix $\Theta$ to extract $I_2$ dimensional node feature. Then, the node feature is gathered according to the hyperedge by using the multiplication of $\mathbf{W}\mathbf{H}^T \in \mathbf{R}^{L \times N}$ to form the hyperedge feature, which can be denoted as $\mathbf{W}\mathbf{H}^T \mathbf{X} \Theta \in \mathbf{R}^{L \times I_2}$. After that, by multiplying matrix $\mathbf{H}$ to aggregate the related hyperedge feature, we can transform the hyperedge feature into output node feature $\mathbf{H}\mathbf{W}\mathbf{H}^T \mathbf{X} \Theta \in \mathbf{R}^{N \times I_2}$.

According to equation (15), the output feature of the $i$th node can be expressed as:

$$\mathbf{Y}_i = \sum_{j=1}^{N} \sum_{k=1}^{L} \mathbf{H}(i,k)\mathbf{H}(j,k)w_k \mathbf{X}_j \Theta /(d_i \delta_i) \quad (16)$$

It can be concluded from equation (16) that information is propagated among those nodes within a common hyperedge that contain high order relationship. Besides, the hyperedge with larger weights deserve more confidence (Bai et al., 2021), which means that labels within the hyperedge are smoother.

Based on the above definitions, the hypergraph convolutional networks with $P$ layers can be built using the following formulation:

$$X^{(l+1)} = \sigma\left(\mathbf{D}_v^{-1/2}\mathbf{H}\mathbf{W}\mathbf{D}_E^{-1}\mathbf{H}^T\mathbf{D}_v^{-1/2}\mathbf{X}^{(l)}\mathbf{\Theta}^{(l)}\right) \quad (17)$$

where $\mathbf{X}^{(l)}$ is the signal of hypergraph at the $l$th layer ($l<P$), $\sigma$ is the non-linear activation function like ReLU, $\mathbf{X}^{(0)} = \mathbf{V}$.

Imbalanced distribution of changed and unchanged samples is very typical in CD problem. Generally, the changed regions often occupy a minority. To solve the sample imbalance problem, in the training process, we adopt the focal loss (Lin et al., 2017), which has been proved its availability in various detection and classification tasks. The loss function in this article can be defined as:

$$L_{HGNN} = \sum_{i=1}^{N_L} -\alpha_t \left(1 - p_t^{(i)}\right)^\gamma \log\left(p_t^{(i)}\right) \quad (18)$$

where $\alpha_t$ is used to control the contribution weight of positive (changed) and negative (unchanged) samples to the total loss. When $\gamma$ is 0, it is equivalent to the cross-entropy loss. $N_L$ is the number of labeled samples. In this article, we set $\alpha_t = 0.2$ and $\gamma = 2$. $p_t^{(i)}$ is the prediction probability of the $i$th labeled sample and it is formulated as:

$$p_t^{(i)} = \begin{cases} p_i, & \text{if } y_i = 1 \text{ (changed)} \\ 1 - p_i, & \text{if } y_i = 0 \text{ (unchanged)} \end{cases} \quad (19)$$

where $p_i$ is the predicted output of the networks.

## 4. Experiments and analysis

In this section, the data sets are firstly illustrated. Then, we provide a brief description of the implementation details and evaluation metrics. Following that, we demonstrate the effectiveness of our proposed method on the optical, SAR and heterogeneous optical-SAR data sets, respectively, by both qualitative and quantitative ways. Finally, some discussion about our method is made in details.

### 4.1 Descriptions of Data Sets

Two openly optical VHR image data sets are employed: the first one is the Sun Yat-Sen University CD data set (SYSU-CD) provided by Shi et al (2021). SYSU-CD contains 20000 pairs of aerial images with the spatial resolution of 0.5m and size of 256×256 pixels. They are collected between the years 2007 and 2014 in Hong Kong. The data set includes the following main types of changes: newly built building, suburban dilation, change of vegetation, road expansion and sea construction. Several patch samples of the first data set are shown in Fig.5. The second one is the WHU-CD data set (Ji et al., 2019), as shown in Fig.6. It contains a pair of aerial image with the spatial resolution of 0.2m, which was taken in the years of 2012 and 2016, respectively. The image size is 15354×32507 pixels and the mainly labeled types of changes are construction and decline of buildings. We split the images into 6000 images pairs with the size of 256×256. Both of above data sets introduce variations derived from the seasonal factors and illumination conditions, which could help develop effective methods that can mitigate the impact of irrelevant changes on real changes.

Two VHR data sets are used to conduct the experiments of CD for SAR images. The first data set (Wuhan-SAR) is a pair of images acquired by TerraSAR-X sensor with HH polarization and 1m/pixel covering a suburban area of Wuhan, China, where the remarkable changes are the construction and demolition of buildings. The image size is 780×870, as shown in the first row of Fig.7. The second data set (Shanghai-SAR) corresponds to an area in Shanghai, China, acquired by Gaofen-3 with the size of 457×553 and the spatial resolution of 1m, as shown in the second row of Fig.7.

Five VHR optical/SAR data sets (Data 1~5) are used to evaluate the efficiency of the proposed method in heterogeneous CD tasks. The first one consists of an optical image acquired by QuickBird in July 2006 and a SAR image captured by TerraSAR-X in July 2007, with the spatial resolution of 0.65m. The mainly changed information is the flooding in Gloucester, England. The second to the fifth data sets all consist of a Google image and a Gaofen-3 SAR image, covering different cities in China, including Shanghai, Zhengzhou, Huizhou and Xiangshui. The spatial resolution is 1m and the main types of changes are construction and decline of buildings, areas of water covering. These five data sets are shown in Fig.8.

### 4.2 Implementation and Evaluation Metrics

The proposed method is implemented via Pytorch framework on a single GeoForce GTX1080Ti GPU. In the feature extracting network training phase, images with labels in the OSCD data set are clipped into 6400 training images of 112×112 pixels with data augmentation, including rotation, flip. The stochastic gradient descent (SGD) with momentum is applied for training. The learning rate is fixed and set to 0.001, meanwhile, the momentum and the weight decay is set to 0.9 and 0.0005, respectively. For each image pair introduced in Section 4.1, we randomly select 5% of the objects under the fine scale parameter as labeled nodes. The number of HGNN layers is set to 2, namely $P=2$ in formula (17). We train the proposed networks for 400 epochs with the dropout rate of 0.5 and the weight decay of 0.0005.

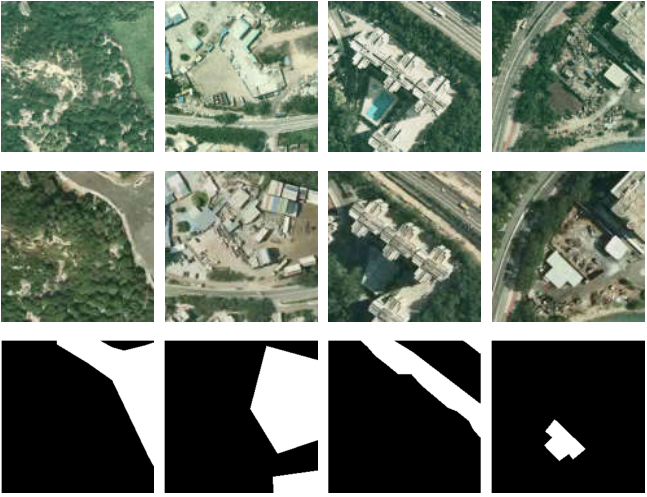

Fig.5. Image patch examples and corresponding reference images of the SYSU-CD data set.

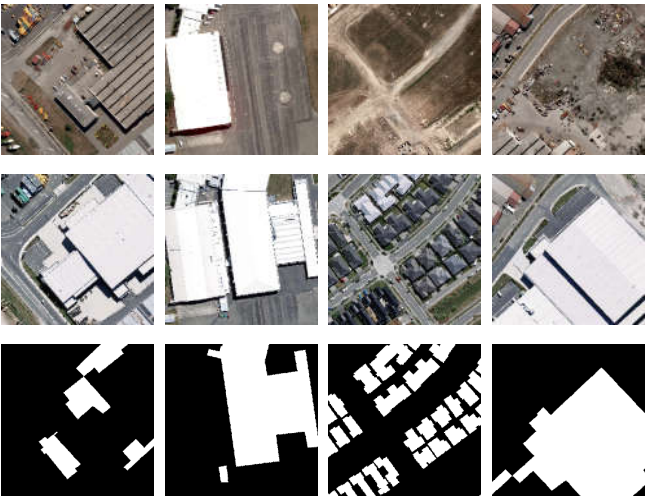

Fig.6. Image patch examples and corresponding reference images of the WHU-CD data set.

To evaluate the performances of the proposed method, four quantitative evaluation indices, false alarm rate (FAR), missed alarm rate (MAR), overall accuracy (OA) and Kappa coefficient (Kappa) are adopted as metrics. The equations of *FAR*, *MAR* and *OA* can be formulated as: *FAR=FP/(FP+TN)*, *MAR=FN/(FN+TP)* and *OA=(TP+TN)/(TP+TN+FP+FN)*,

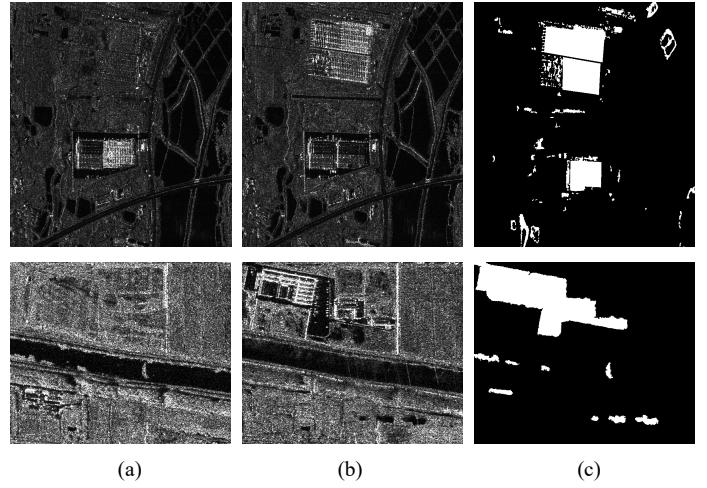

Fig.7. SAR data sets. (a) Image T1. (b) Image T2. (c) Reference change map

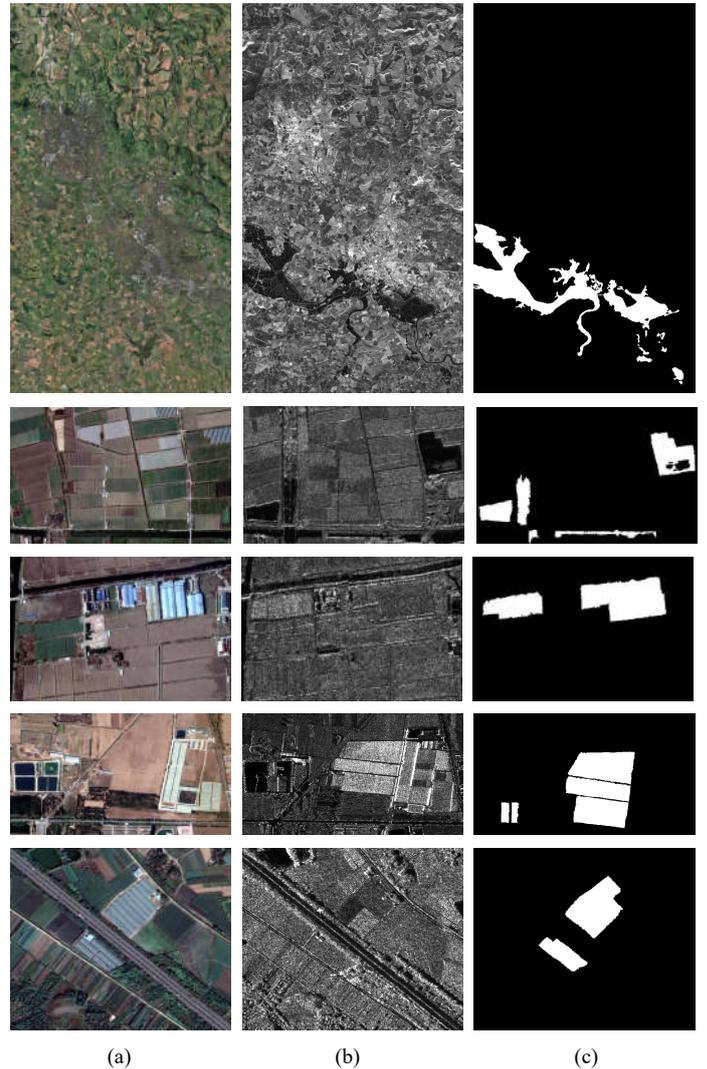

(a) (b) (c)

Fig.8. Heterogeneous optical/SAR data sets. (a) Optical image. (b) SAR image. (c) Reference change map

respectively, where *TP* denotes the number of true positives, *FP* denotes the number of false positives, *TN* denotes the number of true negatives, and *FN* denotes the number of false negatives. *Kappa* is a statistical measure of the consistency between the change map and the reference. It is calculated by:

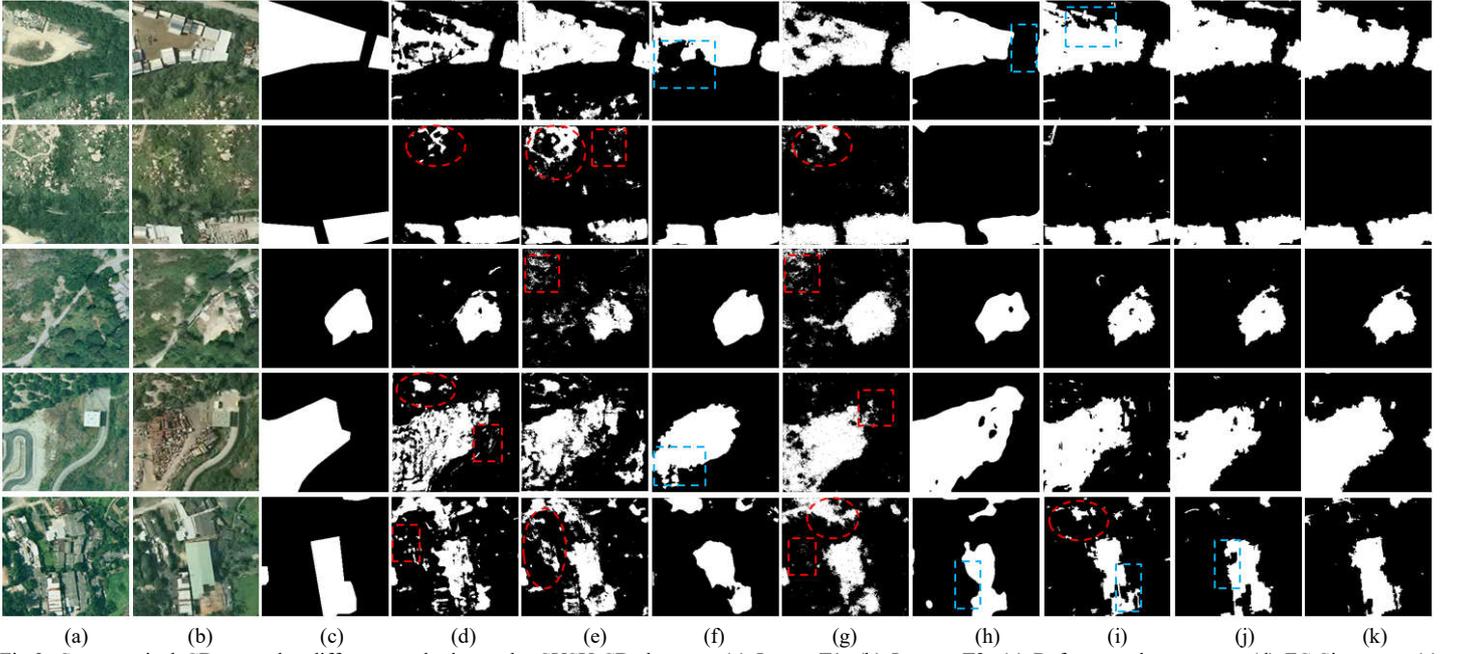

Fig.9. Some typical CD maps by different methods on the SYSU-CD data set. (a) Image T1. (b) Images T2. (c) Reference change map. (d) FC-Siam-con. (e) FC-Siam-diff. (f) Siam-NestedUNet. (g) DSFIN. (h) DSAMNet. (i) GCNCD. (j) MSGCN. (k) DNHGNN

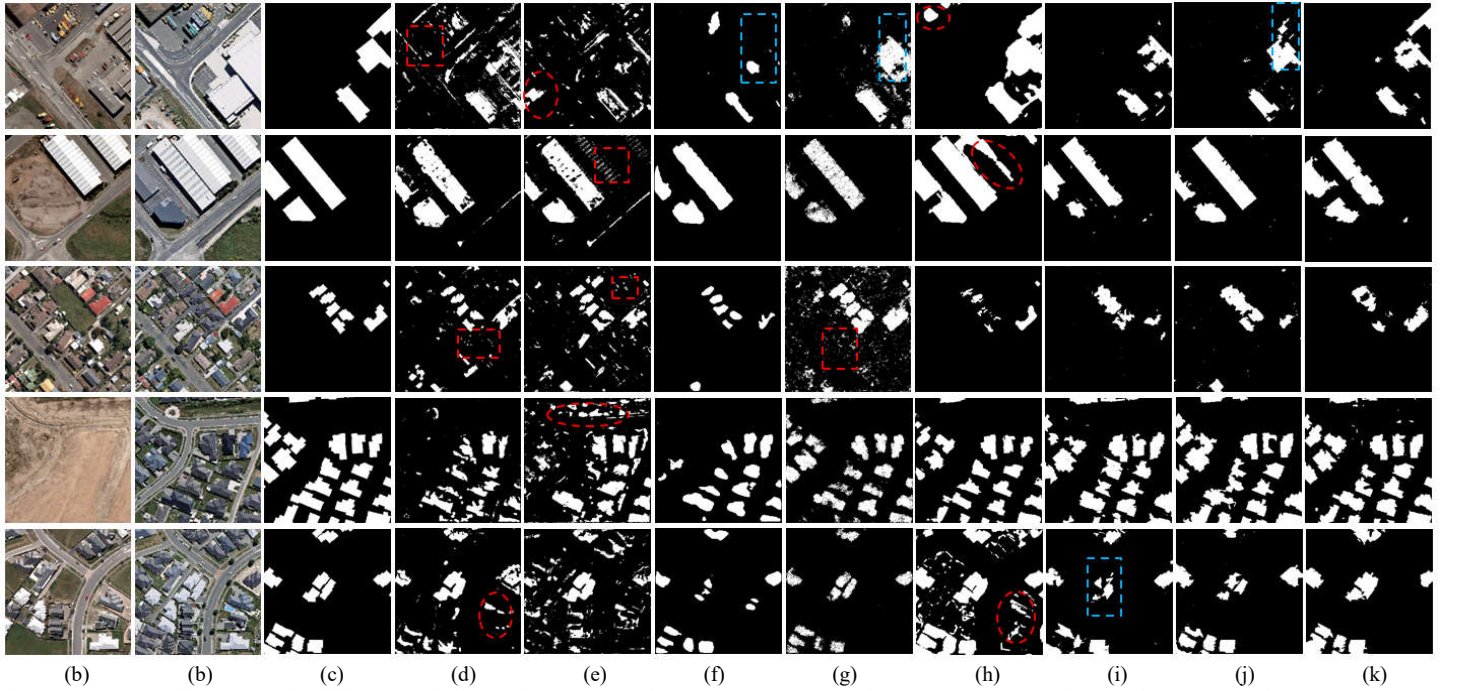

Fig.10. Some typical CD maps by different methods on the WHU-CD data set. (a) Image T1. (b) Images T2. (c) Reference change map. (d) FC-Siam-con. (e) FC-Siam-diff. (f) Siam-NestedUNet. (g) DSFIN. (h) DSAMNet. (i) GCNCD. (j) MSGCN. (k) DNHGNN

$$Kappa = (OA\text{-}PRE)/(1-PRE),$$
$$PRE = \frac{(TP+FN)(TP+FP)+(TN+FP)(TN+FN)}{(TP+TN+FP+FN)^2} \quad (20)$$

### 4.3 Experiments on Optical Images

To verify the effectiveness of the proposed method for optical images, the following seven benchmark methods are employed as competitors on the aforementioned SYSU-CD and WH-CD data sets:

1) FC-Siam-conc: The FC-Siam-conc (Daudt et al, 2018) uses a Siamese encoding stream to extract deep features from bi-temporal images, then, the features are concatenated in the decoding stream.

2) FC-Siam-diff: Different from FC-Siam-conc, FC-Siam-diff (Daudt et al, 2018) uses absolute difference between bitemporal features to decide changed level.

3) Siam-NestedUNet: The Siam-NestedUNet uses a Siamese semantic segmentation network UNet++ to extract features of different resolution, following that, the features are fed into an ensemble channel attention module (Li et al., 2020).

4) DSIFN: A deeply supervised image fusion network (DSIFN) (Zhang and Yue et al., 2020) has been proposed which consists of a shared deep feature network and a difference

Table 1

The quantitative evaluation results of different methods on optical data sets (%)

| Data set | Method | FAR | MAR | OA | Kappa |
|---|---|---|---|---|---|
| SYSU-CD | FC-Siam-conc | 4.57 | 36.29 | 88.91 | 62.30 |
| | FC-Siam-diff | 7.77 | 39.27 | 86.51 | 54.03 |
| | Siam-NestedUnet | 3.26 | 24.41 | 92.12 | 73.47 |
| | DSIFN | 5.20 | 18.75 | 91.79 | 72.72 |
| | DSAMNet | 2.97 | 17.77 | 93.60 | 78.29 |
| | GCNCD | 1.92 | 20.02 | 95.01 | 82.21 |
| | MSGCN | **1.83** | 15.06 | 95.94 | 85.22 |
| | DNHGNN | 2.08 | **10.14** | **96.72** | **87.88** |
| WHU-CD | FC-Siam-conc | 4.31 | 44.79 | 91.03 | 52.46 |
| | FC-Siam-diff | 4.63 | 44.07 | 90.88 | 49.60 |
| | Siam-NestedUnet | **0.89** | 53.55 | 93.07 | 54.54 |
| | DSIFN | 2.39 | 24.78 | 94.49 | 67.83 |
| | DSAMNet | 3.35 | **15.09** | 95.06 | 74.22 |
| | GCNCD | 1.70 | 34.13 | 95.30 | 70.74 |
| | MSGCN | 1.78 | 24.53 | 96.01 | 76.83 |
| | DNHGNN | 1.16 | 18.97 | **97.05** | **82.19** |

Table 2

The quantitative evaluation results of different methods on SAR data sets (%)

| Data set | Method | FAR | MAR | OA | Kappa |
|---|---|---|---|---|---|
| Wuhan-SAR | PCA-Kmeans | 16.71 | 25.97 | 82.48 | 35.35 |
| | ELM | 4.67 | 30.73 | 93.05 | 59.77 |
| | S-PCA-Net | 3.72 | 32.70 | 93.74 | 61.90 |
| | CWNN | 5.70 | 24.42 | 92.62 | 60.36 |
| | CNN | **0.14** | 36.84 | 95.82 | 74.93 |
| | MSGCN | 1.61 | **18.39** | 96.92 | 80.61 |
| | DNHGNN | 1.12 | 19.21 | **97.29** | **82.46** |
| Shanghai-SAR | PCA-Kmeans | 15.91 | 47.45 | 79.97 | 29.34 |
| | ELM | 11.15 | 51.31 | 83.61 | 34.19 |
| | S-PCA-Net | 7.73 | 56.07 | 85.96 | 36.92 |
| | CWNN | 13.24 | 44.65 | 82.66 | 35.54 |
| | CNN | 0.83 | 54.19 | 91.46 | 54.19 |
| | MSGCN | 2.62 | 17.05 | 95.50 | 80.20 |
| | DNHGNN | **0.75** | **10.76** | **97.65** | **89.72** |

discrimination network which utilizes the channel attention module and spatial attention module.

5) DSAMNet: The deeply supervised attention metric-based network (DSAMNet) employs a metric module to learn change maps by means of deep metric learning, in which convolutional block attention modules are integrated to provide more discriminative features (Shi et al., 2021).

6) GCNCD: A network with two GCN layers (Saha et al., 2020) uses several hand-crafted features of objects as node features of the graph model.

7) MSGCN: A multiscale GCN has been proposed which can fuse the outputs of GCN under different segmented scales (Wu et al., 2021).

The first five methods are supervised, thus, we use 12000 pairs of images for training, 4000 pairs for verification, 4000 pairs for testing on the SYSU-CD data set and 3600 pairs for training, 1200 pairs for verification, 1200 pairs for testing on the WHU-CD data set, respectively. The GCNCD and MSGCN are semi-supervised as our method. Thus, to ensure the fairness, we randomly select 5% of the objects under the fine scale as labeled nodes. The two scale parameters of multiscale segmentation are set as 8 and 15 for both data sets.

Some typical results of SYSU-CD and WHU-CD are presented in Fig.9 and Fig.10, respectively.

Intuitively, the results generated by the three GNN-based methods (GCNCD, MSGCN and DNHGNN) can reflect main information of changes. To be specific, one can see that change maps provided by FC-Siam-conc, FC-Siam-diff and DSIFN are affected by significant salt-and-pepper noise where plenty of unchanged pixels are misclassified as changed ones (marked by the red boxes in Fig.9 (d), (e), (g) and Fig.10 (d), (e), (g)), caused by their limited robustness on inevitable misregistration errors and spectral variations in high resolution RS images. Obvious false alarms also occur in the results of these three methods (see the red ellipses in Fig.9 (d), (e), (g) and Fig.10 (d), (e)). In addition, some changed regions in results of DSIFN are not homogeneous enough, such as the first row of Fig.9 (g). The Siam-NestedUnet addresses these issues well by introducing ensemble channel attention module, and thus causes less false alarms. However, the boundaries between changed and unchanged regions are inaccurate in some results, as can be seen in the blue boxes of Fig.9 (f) and Fig.10 (f), the results fail to partition the changed and unchanged regions. The DSAMNet achieves relatively good performances on some image pairs, such as the second row of Fig.9 (h) and the fourth row of Fig.10 (h), but false alarms are still relatively high in some results, such as the red ellipses in Fig.10 (h). Besides, DSAMNet losses some changed regions, which are marked by the blue boxes in Fig.9 (h). GCNCD can obtain results with relatively complete regions. Nevertheless, some boundaries are inaccurate (see the blue boxes in Fig.9 (i) and Fig.10 (i)), due to the discrimination of hand-crafted features is limited. MSGCN seems to obtain similar results as DNHGNN. However, interpreted in detail, the results of DNHGNN are more consistent with the reference maps, especially in some regions with complex structures, DNHGNN achieves higher precision in representing the shapes and boundaries of changed objects, as shown by the blue boxes in Fig.9 (j) and Fig.10 (j).

Table 1 reports the quantitative evaluation results of

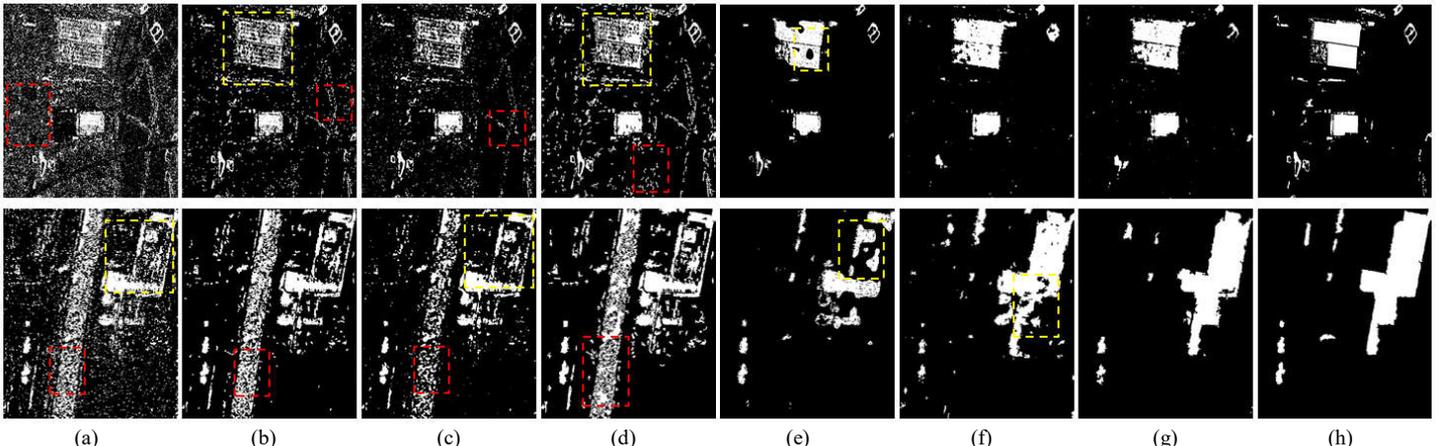

Fig.11. Visual results by different methods on the SAR data sets. (a) PCA-Kmeans. (b) ELM. (c) S-PCA-Net. (d) CWNN. (e) CNN. (f) MSGCN. (g) DNHGNN. (h) Reference change map.

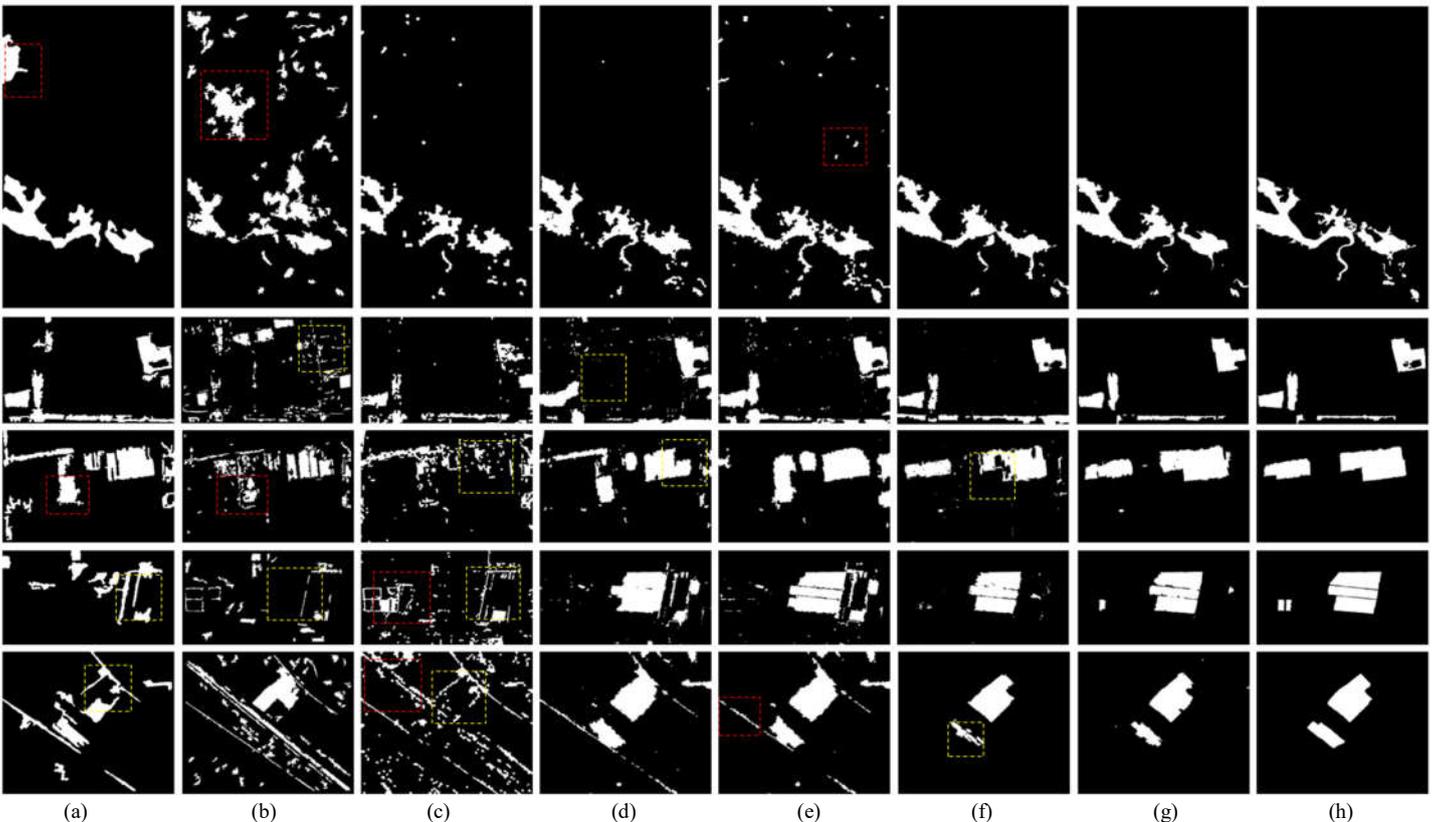

Fig.12. Visual results by different methods on the heterogeneous Optical/SAR data sets. (a) FPMS. (b) CICM. (c) IRG-MCS. (d) SCASC. (e) GIR-MRF. (f) MSGCN. (g) DNHGNN. (h) Reference change map.

different methods. It can be concluded that the quantitative results are consistent with the visual evaluation, and the proposed DNHGNN outperforms all of the compared methods in term of Kappa and OA on both data sets. Besides, it reaches the evidently lowest MAR on SYSU-CD data set and the second lowest on WHU-CD data set. Regarding the FARs, those of DNHGNN are also relatively low. Comparing with other methods, the DNHGNN yields an improvement of at least 4.92%, 0.78%, and 2.66% for MAR, OA and Kappa, respectively, on SYSU-CD data set. The improvement on WHU-CD data set is also evident, with the improving values of at least 1.04% and 5.36% for OA and Kappa, respectively. For SYSU-CD data set, GCNCD and MSGCN achieve slightly lower FARs than that of DNHGNN. However, they yield significantly higher MARs, resulting in loss of some structural information in changed regions as analysis above. On the whole, the DNHGNN can suppress the false alarms and reduce the missed detection, simultaneously. The reasons of this behavior are: 1) the combination of pixel-wise high level features with object-based extraction improve the robustness on the misregistration errors and spectral variations, and this can reduce the false alarms and missed detection; 2) When constructing hypergraph, the utilization of dual neighborhood helps to capture structural information of complex regions, effectively

Table 3
The quantitative evaluation results of different methods on heterogeneous Optical/SAR data sets (%)

| Data set | Method | FAR | MAR | OA | Kappa |
|---|---|---|---|---|---|
| Data 1 | FPMS | 2.55 | 10.28 | 97.01 | 75.73 |
| | CICM | 8.36 | 34.32 | 90.17 | 38.41 |
| | IRG-MCS | 1.18 | 27.74 | 97.21 | 74.36 |
| | SCASC | 1.21 | 17.98 | 97.65 | 80.47 |
| | GIR-MRF | 2.11 | **6.07** | 97.77 | 81.62 |
| | MSGCN | 1.11 | 13.03 | 98.15 | 84.34 |
| | DNHGNN | **0.56** | 13.26 | **98.66** | **88.10** |
| Data 2 | FPMS | 6.57 | 11.24 | 93.03 | 64.96 |
| | CICM | 8.54 | 82.75 | 85.08 | 13.47 |
| | IRG-MCS | 4.00 | 45.50 | 92.43 | 51.19 |
| | SCASC | 6.63 | 30.40 | 91.33 | 53.30 |
| | GIR-MRF | 6.48 | 16.86 | 92.63 | 62.04 |
| | MSGCN | 3.30 | 14.58 | 95.73 | 75.14 |
| | DNHGNN | **1.05** | **9.93** | **98.19** | **88.53** |
| Data 3 | FPMS | 10.27 | 44.02 | 85.66 | 40.30 |
| | CICM | 4.95 | 53.21 | 89.24 | 45.16 |
| | IRG-MCS | 8.93 | 80.23 | 81.82 | 12.14 |
| | SCASC | 6.58 | 47.76 | 88.47 | 45.60 |
| | GIR-MRF | 8.64 | 38.38 | 87.78 | 47.87 |
| | MSGCN | **1.29** | 17.10 | 96.81 | 84.40 |
| | DNHGNN | 1.57 | **4.05** | **98.13** | **91.44** |
| Data 4 | FPMS | 7.96 | 82.12 | 83.86 | 11.76 |
| | CICM | 5.85 | 87.21 | 83.78 | 10.73 |
| | IRG-MCS | 8.79 | 88.12 | 81.56 | 10.12 |
| | SCASC | 4.89 | 14.28 | 94.03 | 73.13 |
| | GIR-MRF | 6.22 | 11.99 | 93.12 | 70.68 |
| | MSGCN | 0.53 | 10.82 | 98.29 | 91.34 |
| | DNHGNN | **0.38** | **6.19** | **98.95** | **94.77** |
| Data 5 | FPMS | 5.24 | 51.11 | 91.73 | 39.43 |
| | CICM | 9.35 | 39.68 | 88.65 | 35.65 |
| | IRG-MCS | 9.92 | 84.25 | 85.47 | 14.10 |
| | SCASC | 3.86 | **3.95** | 96.14 | 74.63 |
| | GIR-MRF | 3.64 | 4.57 | 96.30 | 75.32 |
| | MSGCN | **0.13** | 17.54 | 98.72 | 88.79 |
| | DNHGNN | 0.19 | 9.56 | **99.19** | **93.22** |

and accurately.

**4.4 Experiments on SAR Images**

We evaluate the performance of DNHGNN on two SAR data sets comparing with the following six state-of-the-art CD methods:

1) PCA-Kmeans. A PCA based method (Celik., 2009), in which the K-means approach is used for binary classification.

2) ELM. An extreme learning machine based method for CD in SAR images (Gao et al., 2016).

3) S-PCA-Net. The S-PCA-Net (Wang et al., 2019) introduces the imbalanced learning process into PCA-Net.

4) CWNN. The method uses convolutional wavelet neural network (CWNN) instead of CNN to extract robust features with better noise immunity for SAR image CD (Gao et al., 2019).

5) CNN. The method uses a novel CNN framework without any preprocessing, which can automatically extract the spatial characteristics (Li et al., 2019).

6) MSGCN (Wu et al., 2021).

The scale parameters of multiscale segmentation in MSGCN and DNHGNN are set as 10, 15 for Wuhan-SAR data set and 15, 25 for Shanghai-SAR data set, respectively.

The visual results on the two SAR data sets are shown in Fig.11. As can be seen, many false alarms appear in the results generated by PCA-Kmeans, ELM, S-PCA-Net and CWNN in the both data sets, as the red boxes in Fig.11 (a)~(d). The reason is that these methods are essential pixel-based methods, which may have poor immunity to the severe speckle noise that is commonly prevalent in VHR SAR images. Besides, the homogeneity within the regions of changed buildings is dissatisfied, where the completeness of the buildings can be hardly represented by the results of these four methods (see the yellow boxes in Fig.11 (a)~(d)). In contrast, CNN, MSGCN and DNHGNN can effectively suppress false alarms. However, more changed regions are missed in the results of CNN. For example, in the regions of yellow boxes in Fig.11 (e), the areas of changed buildings are fragmentary. The results of MSGCN lose more structural information compared with those of DNHGNN, as marked by the yellow box in Fig.11 (f).

Table 2 shows the quantitative evaluation results on the two SAR data sets. DNHGNN achieves the best FAR, MAR, OA and Kappa on Shanghai-SAR data set. For Wuhan-SAR data set, the OA and Kappa of DNHGNN are also the best. Although DNHGNN achieve a higher FAR than CNN, the MAR is reduced by a large margin. Due to the complicated noise situation, the performances of the PCA-Kmeans, ELM, S-PCA-Net and CWNN are dissatisfied. Obviously, the FARs of these four methods are much higher than those of others. Although CNN achieves the lowest FAR, it yields significantly higher MAR, which is as high as 36.84%, and this means that many changed pixels are missed in the results of CNN, as our analysis in the aforementioned visual results. Considering the metrics in Table 2 comprehensively, DNHGNN produces the best results on both data sets.

**4.5 Experiments on Heterogeneous Optical/SAR Images**

To evaluate the efficiency of the proposed DNHGNN on heterogeneous optical/SAR image CD tasks, we validate it comparing with the following benchmark methods:

1) FPMS. The fractal projection and Markovian segmentation based method (FPMS) projects the pre-event image to the domain of post-event image by fractal projection.

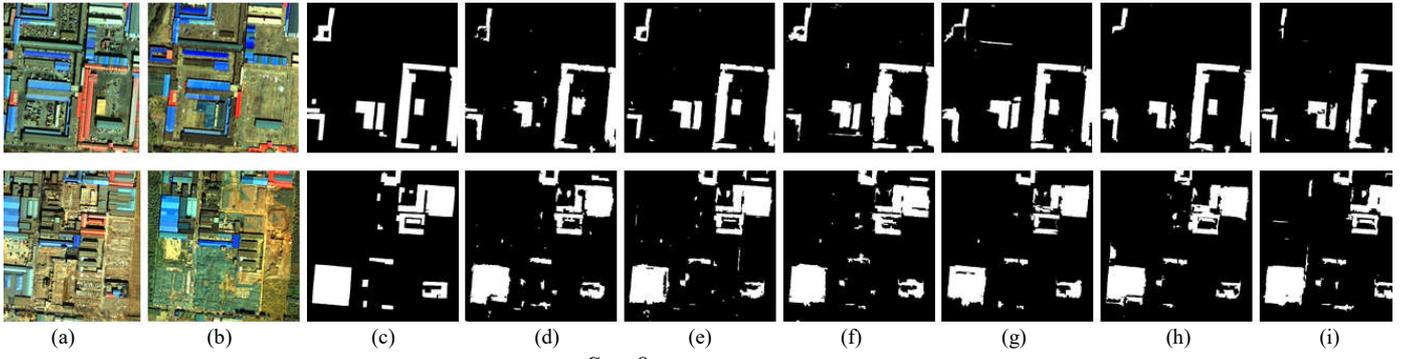

(a) (b) (c) (d) (e) (f) (g) (h) (i)

Fig.13. Two samples of CD results by DNHGNN with fixed $S_1 = 8$ and different $S_2$. (a)Image T1. (b) Image T2. (c) Reference change map. (d) $S_2 = 15$ .(e) $S_2 = 18$ . (f) $S_2 = 21$ . (g) $S_2 = 24$ . (h) $S_2 = 27$ and (i) $S_2 = 30$ .

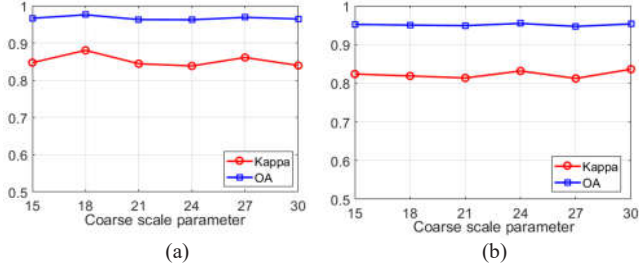

(a) (b)

Fig.14. Kappa and OA under different $S_2$ . (a) Sample 1. (b) Sample 2.

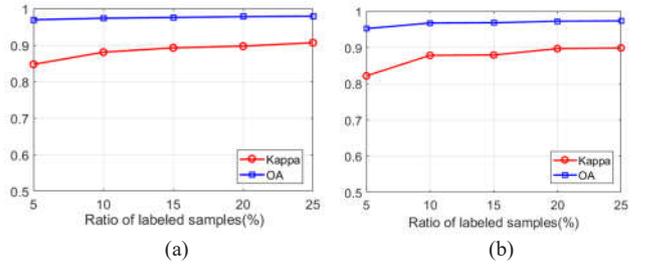

(a) (b)

Fig.15. Kappa and OA with different labeled ratios. (a) Sample 1. (b) Sample 2.

Then, a MRF segmentation model is employed to obtain change maps (Mignotte., 2020).

2) CICM. An concentric circular invariant convolution model (CICM) is proposed to project one image into the imaging modality of the other (Touati., 2019).

3) IRG-MCS (Sun et al., 2021). The authors define an iterative robust similarity graph to measure changed degree.

4) SCASC. An unsupervised method that uses the regression model with sparse constrained adaptive structure consistency (Sun et al., 2022).

5) GIR-MRF. A structured graph learning based method, which first learns a robust graph to capture the local and global structure information of image, and then projects the graph to domain of the other image to complete the image regression (Sun et al., 2022).

6) MSGCN (Wu, et al., 2021).

The scale parameters of multiscale segmentation in MSGCN and DNHGNN are set as 25, 50 for the first data set and 12, 30 for the others.

Fig.12 shows the changed maps of all comparing methods on the heterogeneous optical/SAR data sets. It can be observed that the performances of FPMS, CICM, and IRG-MCS are unsatisfactory, as they not only cause many false alarms (marked by red boxes in Fig.12 (a)~(c)), but also miss some mainly changed regions, such as those shown by the yellow boxes in Fig.12 (a)~(c). SCASC seem to achieve better performances than those of the above three ones. Nevertheless, missing of changed regions still exist, as can be seen in the yellow boxes in Fig.12 (d). Intuitively, GIR-MRF and MSGCN can capture the mainly changed information as DNGHNN does. However, interpreted in details, GIR-MRF causes more false alarms than DNGHNN (see the red boxes in Fig.12 (e)). Compared with DNGHNN, MSGCN fails to accurately capture some shapes of changed regions, such as those marked by the yellow boxes in Fig.12 (f).

The quantitative evaluation results on the two heterogeneous data sets are listed in Table 3. DNHGNN outperforms other methods significantly in terms of OA and Kappa. As it is consistent with the visual comparison of Fig.12, the relatively low MAR and FAR mean that DNHGNN can effectively suppress the false alarms and avoid the missed detection, simultaneously. On the whole, the proposed DNHGNN outperforms the benchmark methods and the reasons of this behavior may be: 1) Different from the methods based on regression or feature projection, DNHGNN treats CD as a classification task and obtain features from the same feature space, which may avoid the influence of heterogeneity between images. 2) A small amount of labeled samples (5%) supervise the hypergraphs which contain structural information to identify the labels of nodes.

### 4.6 Discussion

In the following, the robustness to the scale parameters, Influence of the ratio of labeled samples will be discussed in detail.

*(1) Robustness to the scale parameters*

In order to capture more comprehensive information of the ground objects and reduce the number of nodes, we segment the input image pair under one fine scale parameter $S_1$ and one coarse $S_2$, respectively. Each object under $S_1$ is treated as a node and the father-child relationships between objects under $S_1$ and $S_2$ are used to construct hypergraphs. Thus, the scale parameters are pivotal to the performance of DNHGNN.

To ensure the homogeneity of each objects, we set $S_1$ to a relatively small value to guarantee moderate over-segmentation. Nevertheless, $S_2$ needs to be further considered. As an object under $S_2$ are merged by several objects under $S_1$, different $S_2$ can obtain distinctly structural information. To be specific, when fixing the fine scale parameter, a larger $S_2$ would result in more nodes within a hyperedge, and this means that usable information within a hyperedge increase, but may lead to redundancy. Consequently, in this section, for each image pair, we fix $S_1$ and set $S_2$ to different values to discuss the performance.

Fig.13 shows two samples of optical CD results with $S_1 = 8$ and different $S_2$, in steps of 3. As can be seen in the first row of Fig.13, when $S_2$ increases from 15 to 18, the changed building in the up-left region is detected more precisely, while $S_2$ increases from 21 to 24, the detected structure of this building is not integrated again due to the utilization of some redundant information. This phenomenon also occurs in the second row of Fig.13. On the whole, all results shown in Fig.13 can capture main changed information with some discrepancy in complex regions. Meanwhile, all these results can effectively suppress false alarms, as we can see that, a small number of unchanged pixels are misclassified as changed ones.

The corresponding OA and Kappa are displayed in Fig.14. It can be observed that both OA and Kappa maintain at relatively large values and the discrepancy is not evident. Although the best $S_2$ is hard to determine, the performances under all $S_2 \in [15, 30]$ are acceptable. It means that we can set $S_2$ in a relatively large range. In another word, the method has good robustness to the scale parameters.

*(2) Influence of the ratio of labeled samples*

Unlike other supervised methods which use some individual pairs of images as training data and other pairs as testing data, the semi-supervised DNHGNN performs training with a few labeled superpixels (objects) on each pair. Therefore, the ratio of labeled samples would unavoidably influence the performance and this influence should be investigated. To this end, we vary the ratio of labeled samples from 5% to 25% in steps of 5% and report the OA and Kappa coefficient on the two samples of Fig.13, as shown in Fig.15. We can make the observation from Fig.15 that the performance on both data sets can be improved by increasing the ratio of labeled examples. The improvement is more remarkable when labeled ratio is low. It is noteworthy that the proposed DNHGNN can yield relatively high precision accuracy even though the labeled ratio is low, as the experiments shown in Section 4.3~Section 4.5, where DNHGNN achieves promising performance with the labeled ratio of only 5%..

## 5. Conclusions

In this article, a semi-supervised hypergrpah neural network is proposed for change detection, which can be applied for both homogeneous and heterogeneous remote sensing images. The main idea of the presented framework is to model the high order relationships commonly existing in VHR images using hypergraph and to propagate information through the defined hypergraph convolution. To reduce the number of nodes and obtain object-wise feature with semantic information, the input image pair is firstly segmented through FNEA with one fine scale parameter and one coarse, respectively. Treating each object under fine scale parameter as a node, the node features can be obtain by combining the segmentation with a pre-trained U-net. The hyperedges are constructed based on a defined dual neighborhood which uses not only spatial adjacent relationships under the fine scale but also the father-child relationships between the two scales. After that, the constructed hyperedges can represent complex structure more comprehensively. With the help of the structural information, the hypergraph convolution operations can accurately propagate the label information from labeled nodes to unlabeled ones. The experimental results have demonstrated the superiority of the proposed DNHGNN against some popular methods. Our future work is to explore the possibility of distinguishing different kinds of changes and extend the framework to time-series images.

**Acknowledgments:**

This study is supported by National Natural Science Foundation of China under Grant 62001482.